\documentclass[10pt,twocolumn,letterpaper]{article}

\usepackage{iccv}
\usepackage{times}
\usepackage{epsfig}
\usepackage{graphicx}
\usepackage{amsmath}
\usepackage{amssymb}
\usepackage{multirow}
\usepackage{booktabs}
\usepackage{caption}
\usepackage{array}
\usepackage{adjustbox}

\usepackage[pagebackref=true,breaklinks=true,letterpaper=true,colorlinks,bookmarks=false]{hyperref}

\iccvfinalcopy 

\def\iccvPaperID{2152} 

\ificcvfinal\fi
\begin{document}

\title{Mimicking the In-Camera Color Pipeline for Camera-Aware Object Compositing}

\author{Jun Gao$^{1,3}$\thanks{This work was done when Jun Gao was an intern at Microsoft Research.}\hspace{1cm} Xiao Li$^{2,5}$ \hspace{1cm} Liwei Wang$^{6,7}$ \hspace{1cm} Sanja Fidler$^{1,3,4}$ \hspace{1cm} Stephen Lin$^{5}$\\
$^{1}$University of Toronto\\
$^{2}$University of Science and Technology of China \hspace{0.5cm} $^{3}$Vector Institute \hspace{0.5cm} $^{4}$NVIDIA \hspace{0.5cm} $^{5}$Microsoft Research\\
$^{6}$Key Laboratory of Machine Perception, MOE, School of EECS, Peking University\\ $^{7}$Center for Data Science, Peking University, Beijing Institute of Big Data Research\\
{\tt\small \{jungao,fidler\}@cs.toronto.edu, pableetoli@gmail.com, wanglw@cis.pku.edu.cn, stevelin@microsoft.com}
}

\maketitle

\begin{abstract}
   We present a method for compositing virtual objects into a photograph such that the object colors appear to have been processed by the photo's camera imaging pipeline. Compositing in such a camera-aware manner is essential for high realism, and it requires the color transformation in the photo's pipeline to be inferred, which is challenging due to the inherent one-to-many mapping that exists from a scene to a photo. To address this problem for the case of a single photo taken from an unknown camera, we propose a dual-learning approach in which the reverse color transformation (from the photo to the scene) is jointly estimated. 
   Learning of the reverse transformation is used to facilitate learning of the forward mapping, by enforcing cycle consistency of the two processes. We additionally employ a feature sharing schema to extract evidence from the target photo in the reverse mapping to guide the forward color transformation. Our dual-learning approach achieves object compositing results that surpass those of alternative techniques. 
\end{abstract}
\section{Introduction}
\label{sec:intro}
\begin{figure}
	\centering
	\includegraphics[height=2.7cm]{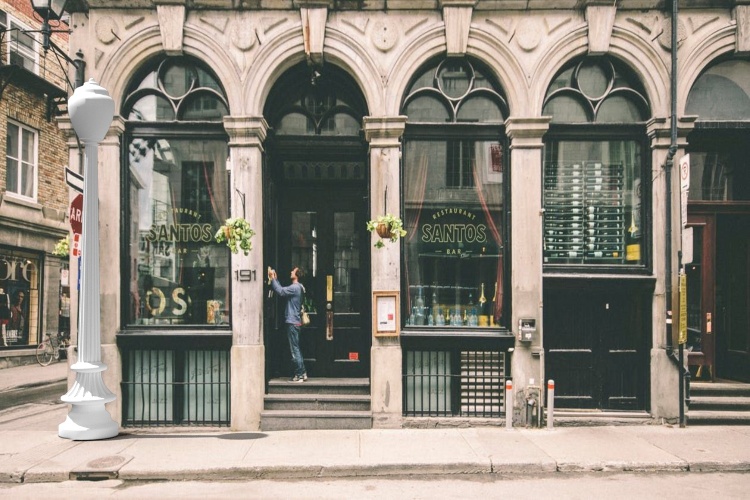}
	\includegraphics[height=2.7cm]{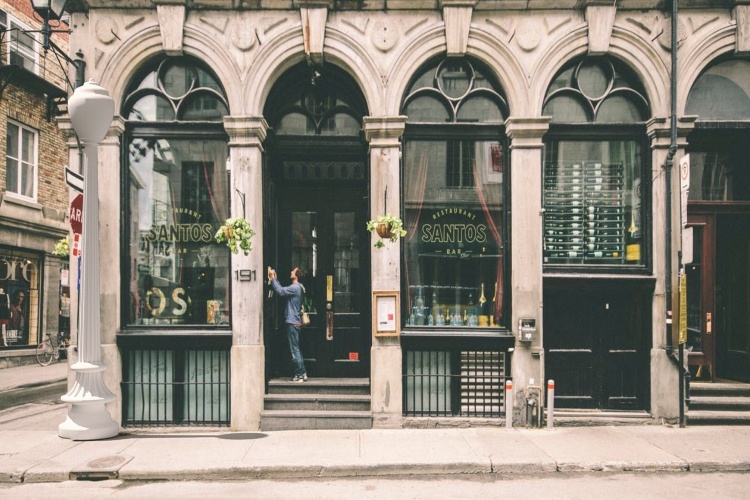}
	\vspace{-5pt}
	\caption{{\bf Compositing of a virtual streetlamp.} Left: Directly compositing the object model. Right: Compositing the object model via the proposed color translation network.}
	\vspace{-10pt}
	\label{fig:example}
\end{figure}

Compositing virtual objects into real photographs, such as adding a streetlamp in front of a building, is a common feature in interactive applications such as augmented reality.
While this can be done with current computer vision technology, making the composited object look realistic remains a challenge. Even with a highly detailed object model and known illumination conditions, the object's appearance in a photo can appear unnatural because its colors do not conform with the rest of the scene, as shown in Figure~\ref{fig:example}(left). The colors in a photograph are a result not only of the scene content, but also of the imaging pipeline in the camera, which may include color filters, white balancing, and dynamic range compression. For attaining high realism, a composited virtual object needs to undergo the same color transformations as the rest of the image, so that it can blend seamlessly into the photo, as exemplified in Figure~\ref{fig:example}(right). 

If the camera that took the photo is available, one could capture a collection of aligned RAW\footnote{RAW: minimally processed image data from the sensor of a camera.}-JPEG image pairs and train a network to map RAW to JPEG~\cite{nam2017modelling}, as the virtual object is in the RAW domain. However, for broad applicability, it is desirable to composite virtual objects into photos taken by unknown cameras for which we have no training data. Since there exist many possible color transformations from RAW to JPEG due to the broad diversity of camera imaging pipelines, finding the specific one that generated a given photo, without having the camera, is a non-trivial problem. This is the core challenge addressed in our work.

To address this issue, we present a deep dual-learning approach in which images are bidirectionally transformed between the JPEG colors produced by the imaging device and a canonical RAW color space in which the virtual object is represented. In particular, to aid in learning the RAW-to-JPEG transformation (primal network), we simultaneously learn a JPEG-to-RAW transformation (dual network), which is more practical to learn, since it represents a one-to-one mapping (i.e., a JPEG image captures a specific scene). Furthermore, there exist many objects such as grass, sky, and human skin that span a limited range of colors in natural scenes and thus provide strong constraints on this mapping. To facilitate learning of the coupled networks, we employ cycle consistency~\cite{zhu2017unpaired,he2016dual}, in which translating JPEG to RAW and back to JPEG should be an identity mapping. However, after training, as the primal network itself is still a deterministic function, it can only represent a one-to-one mapping. We thus propose a feature sharing scheme where features extracted by the dual network from the JPEG are passed to the primal network to guide the RAW-to-JPEG transformation. 

Given an object model and a target JPEG image at test time, our system feeds the JPEG into the dual network to obtain an image in the canonical RAW color space and the corresponding shared features. The object is rendered into the image, which, together with the shared features, is then input into the primal network to generate the compositing result. Although there exist many camera-dependent ways in which a RAW image can be translated to JPEG, the color translation in the primal network is determined by the original JPEG image through the neural features extracted and shared by the dual network. Done this way, the primal network generates a result that mimics the color imaging pipeline of the JPEG input, without needing training data from its camera.

A user study shows that our proposed approach leads to compositing results that are perceptually more coherent than from common baseline techniques. An empirical examination of different variants of this approach is presented as well.


\section{Related Work}
\paragraph{Image Pipeline Modeling}
Physics-based computer vision methods such as shape-from-shading require measurements of scene radiance that are physically accurate. Towards obtaining accurate measurements from photographs, the imaging pipeline of cameras has been modeled and used to undo the effects of in-camera processing. Many techniques have been proposed for modeling a particular component of an imaging pipeline, such as tone mapping \cite{chakrabarti2014modeling,kim2008robust,lin2004radiometric} or white balancing \cite{barron2017fast,hu2017,shi2016deep}. More comprehensive are works that aim to model the sequence of processing operations that occur within an imaging device \cite{chakrabarti2009empirical,kim2012new}. Recently, a deep neural network was presented for modeling the scene-dependent color processing of a given camera, where RAW-JPEG image pairs are captured from the camera for training \cite{nam2017modelling}. In our work, we utilize this deep network for modeling color transformations in the imaging pipeline, but infer the model using only a single photograph from an unknown camera. This inference from a single image is made possible through the use of contextual color priors on common scene objects and our proposed dual-learning approach with a feature sharing schema.

\vspace{-15pt}
\paragraph{Image Compositing}
For increasing the realism of objects composited into photographs, methods have been presented for estimating scene illumination \cite{karsch2011rendering,knorr2014real} and for recovering camera distortions such as those resulting from sensor noise and motion blur \cite{fischer2006enhanced}, or caused by the camera's lens and rolling shutter \cite{klein2010simulating}. In contrast to these previous techniques, our work seeks to heighten realism by estimating and applying the in-camera color processing to composited objects, and thus is complementary to this prior research. Moreover, unlike the methods that model imaging distortions \cite{fischer2006enhanced,klein2010simulating}, which require access to the camera for calibrating these effects, our method is specifically developed not to need the camera at hand, so that it can be applied to arbitrary images.

\vspace{-15pt}
\paragraph{Image-to-Image Translation}
Many image processing problems can be viewed as translating an input image into an output image that exhibits a different representation of the scene. A general framework for this translation problem was introduced using a Generative Adversarial Network (GAN) that learns this mapping from a training set of aligned image pairs from the two domains \cite{isola2017image}. To relax the requirement of paired training data, recent methods have exploited the duality in the image translation problem by jointly learning an additional GAN that maps images from the output domain to the input domain while enforcing a cycle-consistency constraint in which an image mapped from the input domain to the output domain and then back to the input domain should yield the original input \cite{zhu2017unpaired,liu2017unsupervised,kim2017learning}. Through this coupling of GANs, the training data need not be paired, but rather it is sufficient to have independent sets of images in each of the two domains.


Modeled by a deterministic network, the translation learned in these prior works is a {\it one-to-one} mapping, where an image in one domain corresponds to a specific image in the other domain, and vice versa. By contrast, our work deals with a {\it one-to-many} mapping (RAW-to-JPEG) that arises from the differences in imaging pipelines among different in-camera processes, and we focus on how to determine the correct transformation for the {\it one-to-many} mapping.

\section{Dual Learning for Object Compositing}
\begin{figure}
	\centering
	\includegraphics[height=4.3cm]{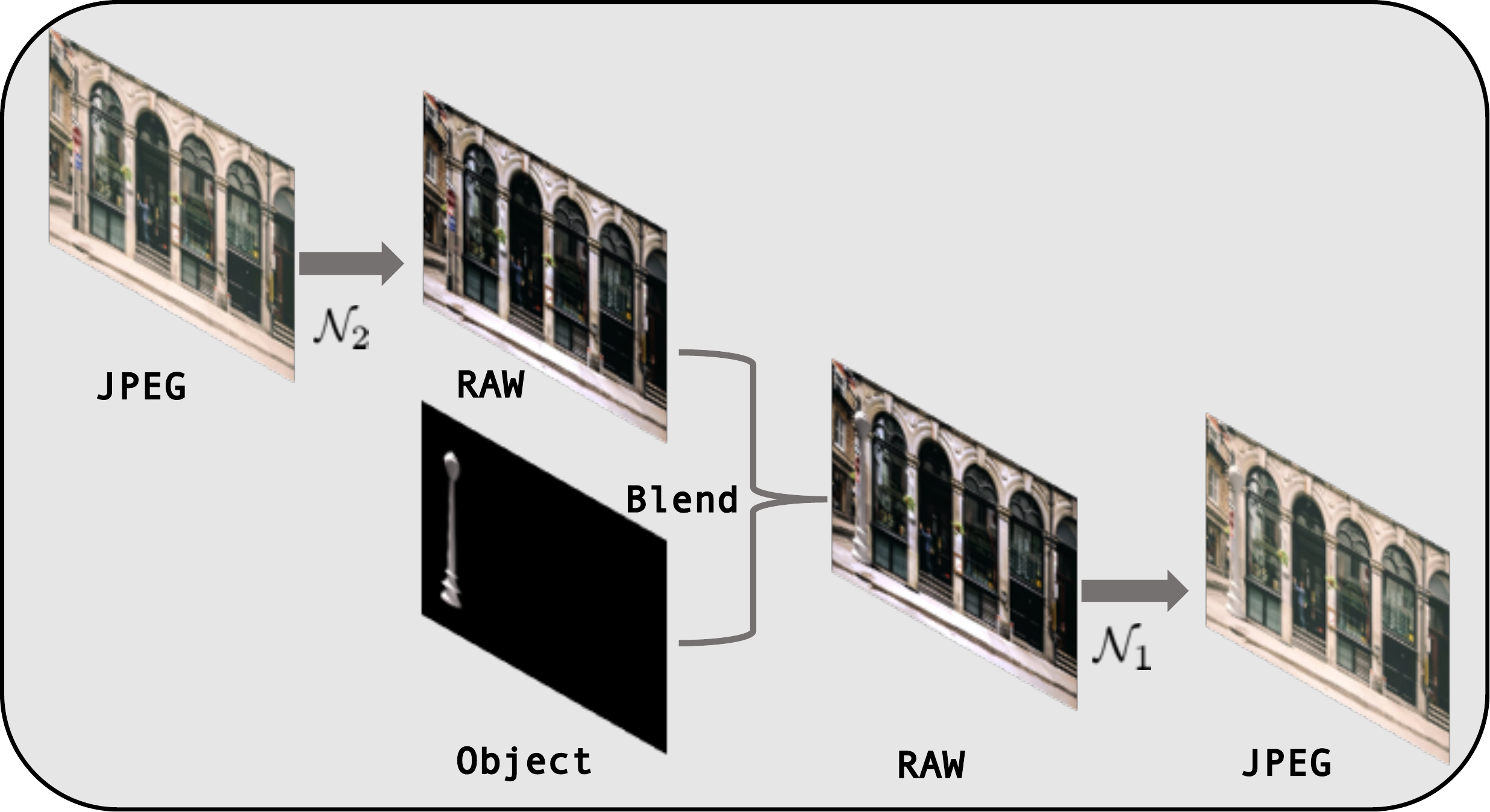}
	\vspace{-8pt}
	\caption{Overview of object compositing using the learned primal network $\mathcal{N}_1$ and dual network $\mathcal{N}_2$.}
	\vspace{-5pt}
	\label{fig:pipeline}
\end{figure}

Our approach proceeds as follows. We first feed the target JPEG image into the dual network (JPEG-to-RAW) $\mathcal{N}_2$ in order to translate it into a demosaiced RAW image in a canonical color space. The virtual object, also represented in the canonical space, is then rendered under the estimated lighting conditions and then composited into the RAW image. Here, we utilize an existing technique for illumination estimation \cite{holdgeoffroy-cvpr-17} and focus on the composition task. The compositing result is then obtained by passing the composited RAW image through the primal network (RAW-to-JPEG)  $\mathcal{N}_1$. An overview of this process is illustrated in Figure~\ref{fig:pipeline}.

We first introduce the canonical color space and a method for transforming a specific camera's RAW image colors to this space in Sec.~\ref{sec:canonical}. Then we present the primal and dual networks and their training algorithms in Sec.~\ref{sec:img_translation}. The object compositing method based on these networks is described in Sec.~\ref{sec:insert_object}.

\subsection{Canonical Color Space}
\label{sec:canonical}
Our system translates image colors between the input JPEG image and a canonical color space in which a virtual object can be represented. As part of learning color translations to and from this canonical space, we capture RAW images from multiple cameras, and transform the camera-dependent RAW image colors to the canonical space through color camera calibration.

Color calibration of cameras is conventionally performed using a ColorChecker chart, which contains patches of known colors 
\cite{pascale2006rgb}. For a RAW image taken of a ColorChecker chart, we thus know the RAW image colors of the patches and their corresponding colors in various color spaces. 
In this paper, we choose linear sRGB as the canonical color space.
With the correspondence among RAW and sRGB colors, the RAW-to-sRGB color transformation can be expressed as follows:
\vspace{-5pt}
\begin{eqnarray}
I_c = F(I_r),
\vspace{-8pt}
\end{eqnarray}
where $I_c$ denotes colors in linear sRGB, $I_r$ represents corresponding colors in the RAW image, and $F$ is a mapping function. To model $F$, we utilize a linear transformation $T_{3\times4}$, which has been found to give the best performance among several candidate models for color calibration \cite{nguyen2014raw}. The mapping is optimized via least squares fitting after subtracting the black level value from the RAW color values.
\vspace{-10pt}


\subsection{Image Translation}
\label{sec:img_translation}
Mapping of an image between the canonical RAW and the photo's JPEG domain can be modeled as an image translation problem, which has been widely studied for applications including image colorization and super-resolution \cite{zhu2017unpaired,ledig2016photo,zhang2016colorful}. For the two mappings, we train two networks denoted as $\mathcal{N}_1$ and $\mathcal{N}_2$, the first for RAW-to-JPEG prediction and the other for estimating JPEG-to-RAW.  
\begin{figure*}
	\centering
	\includegraphics[height=6cm]{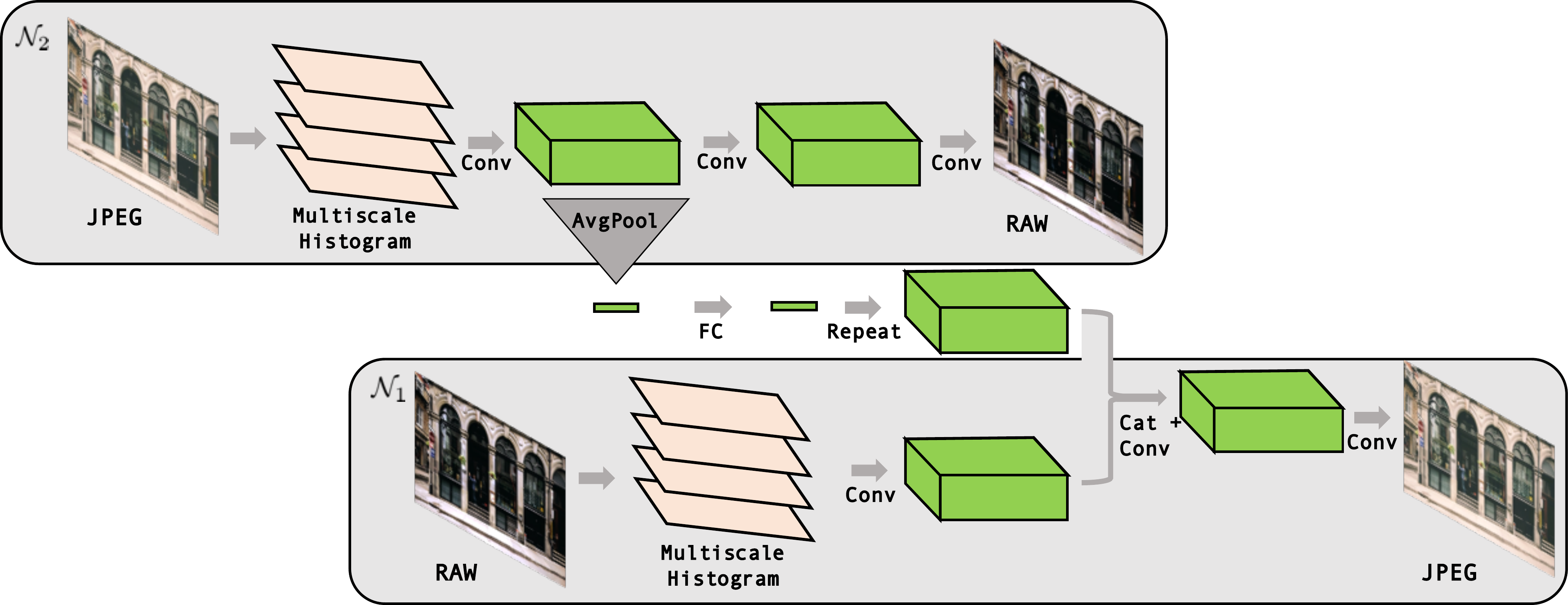}
	\vspace{-7pt}
	\caption{Network Architecture. For $\mathcal{N}_1$ and $\mathcal{N}_2$, we first extract multi-scale histogram features from the input image, which are further processed with three convolutional layers to predict the image in the target domain. We use Average Pooling to extract the shared features from $\mathcal{N}_2$ and propagate them to $\mathcal{N}_1$ via a single Fully-Connected layer followed by repetition.}
	\vspace{-15pt}
	\label{fig:network}
\end{figure*}

\begin{table}
    \begin{center}
    {
        \caption{Network Configuration. Image size of h*w.}
        \label{tbl:network-config}
        \vspace{-10pt}
        \begin{adjustbox}{width=0.48\textwidth}
        \begin{tabular}{lc|lc}
            \toprule
            \multicolumn{2}{c|}{ $\mathcal{N}_1$}&\multicolumn{2}{c}{$\mathcal{N}_2$}\\
            Layer & Output Size & Layer & Output Size\\
            \hline
            Hist & (3$\times6\times4$+3)*h*w & Hist & (3$\times6\times4$+3)*h*w \\
            Conv1 & 128*h*w & Conv1 & 128*h*w\\
            Conv2 & 128*h*w & Conv2 & 128*h*w\\
            Conv3 & 3*h*w & Conv3 & 3*h*w\\
            FC  &  128 & & \\
            \bottomrule
        \end{tabular}
        \end{adjustbox}
        }
    \end{center}
\end{table}
\subsubsection{Network Architecture} 
The structure of our network is illustrated in Figure~\ref{fig:network}, with the network configuration details given in Table~\ref{tbl:network-config}. Our networks $\mathcal{N}_1$ and $\mathcal{N}_2$ are adopted from the Multiscale Learnable Histogram network in \cite{nam2017modelling}, which achieves state-of-the-art performance on radiometric calibration. The networks first extract color histogram features from the input image with learnable bin centers and widths. The histograms are then computed within a multi-scale pyramid, allowing global and local context to be extracted and combined for each pixel. The stacked histograms and images are fed into a three-layer convolutional neural network to predict the output image. Additionally, $\mathcal{N}_2$ produces a feature vector that encodes the global color transformation properties of the JPEG photo and is forwarded to $\mathcal{N}_1$ to aid in predicting the final JPEG image. The design of this feature sharing scheme from $\mathcal{N}_2$ to $\mathcal{N}_1$ will be described in Sec.~\ref{sec:feature_sharing}.

Note that other image translation models could potentially be used for $\mathcal{N}_1$ and $\mathcal{N}_2$. A model should ideally satisfy two properties: (1) the network should be able to account for the high-level global semantic content of the image so that objects which can constrain the mapping (i.e., those with a restricted range of natural colors) are all jointly considered in determining a color transformation; (2) it should be able to extract low-level local color information that reflects the properties of the color transformations in the imaging pipeline. We found multi-scale histogram pyramids to be effective at capturing these two types of information in images. We also tried deep encoder-decoder networks with skip connections \cite{ronneberger2015u}. This yielded worse results, likely because deeper networks are better at extracting high-level semantics but discard low-level information. 
\vspace{-8pt}
\subsubsection{Feature Sharing}
\label{sec:feature_sharing}
As mentioned in the Sec.~\ref{sec:intro}, RAW-to-JPEG color translation is dependent on the imaging pipeline. It thus requires information related to the pipeline's color processing that produced the input JPEG photo. 
For this, we extract features from a hidden layer in $\mathcal{N}_2$ and share them with $\mathcal{N}_1$. Hidden layers in $\mathcal{N}_2$ can provide clues on the color processing, as $\mathcal{N}_2$ seeks to separate the JPEG color processing from the intrinsic colors of objects in the scene. Let us denote the output of the hidden layer as $l$. When passing $l$ from $\mathcal{N}_2$ to $\mathcal{N}_1$, we transform it by a function $f$:
\vspace{-5pt}
\begin{eqnarray}
l^{\prime} = f(l),
\vspace{-5pt}
\end{eqnarray}
where $l^{\prime}$ denotes the features received by $\mathcal{N}_1$. For $f$, we employ average pooling followed by a fully connected layer. Effective in extracting global features \cite{qi2017pointnet}, average pooling facilitates inference of global color transformations. The fully connected layer learns to adapt the features so that they become compatible with the feature space of $\mathcal{N}_1$. In $\mathcal{N}_1$, $l^{\prime}$ is repeated across the spatial dimension to be consistent in size with the feature map in $\mathcal{N}_1$, so that it can be easily concatenated with the feature map and processed by the convolutional layers.

Along the processing hierarchy of our JPEG-to-RAW network $\mathcal{N}_2$, the feature maps should provide an image representation increasingly sensitive to the canonical RAW colors. In contrast to the later layers, the earlier layers more closely represent the JPEG coloring effects of the imaging pipeline. We thus choose the feature map after the first convolutional layer in $\mathcal{N}_2$ as $l$. Related concepts have been used for style transfer, where style information is extracted from earlier layers while content-related features are obtained from later layers \cite{gatys2016image,luan2017deep}. These shared features are inserted into $\mathcal{N}_1$ at its first convolutional layer, so that this information can be accounted for throughout the subsequent layers. The whole model can be expressed as:
\vspace{-5pt}
\begin{eqnarray}
\vspace{-5pt}
I_{\mathrm{RAW}}^{\prime}, l &=& \mathcal{N}_2(I_{\mathrm{JPEG}}), \\
l^{\prime} &=& f(l),\\
I_{\mathrm{JPEG}}^{\prime} &=& \mathcal{N}_1(I_{\mathrm{RAW}}, l^{\prime}),
\vspace{-5pt}
\end{eqnarray}
where $I_{\mathrm{RAW}}^{\prime}$ and $I_{\mathrm{JPEG}}^{\prime}$ denote the predicted RAW and JPEG images, respectively, and $I_{\mathrm{RAW}}$ and $I_{\mathrm{JPEG}}$ are the input RAW and JPEG images.
\subsubsection{Training Loss}
To optimize the network parameters, the most straightforward loss function is the mean-squared error between the target images and predicted images of the two networks:
\vspace{-5pt}
\begin{eqnarray}
L_1=\parallel I_{\mathrm{JPEG}} - I_{\mathrm{JPEG}}^{\prime}\parallel_2 + \parallel I_{\mathrm{RAW}} - I_{\mathrm{RAW}}^{\prime}\parallel_2.
\vspace{-5pt}
\end{eqnarray}

However, during inference time, error will accumulate along the processing hierarchy as the JPEG image passes through $\mathcal{N}_2$ and then through $\mathcal{N}_1$. To reduce such error, we encourage cycle consistency \cite{zhu2017unpaired,he2016dual}:
\vspace{-5pt}
\begin{eqnarray}
\mathcal{N}_1\big(\mathcal{N}_2(I_{\mathrm{JPEG}})\big)=I_{\mathrm{JPEG}},
\vspace{-5pt}
\end{eqnarray}
where a JPEG image passed through the JPEG-to-RAW and RAW-to-JPEG networks should yield a predicted image identical to the original JPEG input. This is done by adding the following term to the loss function:
\begin{eqnarray}
L_2 &=& \parallel I_{\mathrm{JPEG}} - \mathcal{N}_1\big(\mathcal{N}_2(I_{\mathrm{JPEG}})\big)\parallel_2.
\end{eqnarray}
A hyperparameter $\lambda$ is introduced to balance the reconstruction loss and the cycle consistency constraint, giving us the overall loss function:
\vspace{-3pt}
\begin{eqnarray}
L &=& \parallel I_{\mathrm{JPEG}} - \mathcal{N}_1(I_{\mathrm{RAW}})\parallel_2 \\
&+&\parallel I_{\mathrm{RAW}} - \mathcal{N}_2(I_{\mathrm{JPEG}})\parallel_2\\
&+& \lambda \parallel I_{\mathrm{JPEG}} - \mathcal{N}_1\big(\mathcal{N}_2(I_{\mathrm{JPEG}})\big)\parallel_2.
\end{eqnarray}

\subsection{Object Compositing}
\label{sec:insert_object}
To composite a synthetic object into a JPEG photo, we first render the object with lighting estimated using the online demo\footnote{http://rachmaninoff.gel.ulaval.ca:8000/} provided by \cite{holdgeoffroy-cvpr-17}. Its RGB values $R$ in the canonical color space and image mask $M$ are obtained with the Blender renderer\footnote{https://www.blender.org/}. At the same time, we also feed the JPEG photo into $\mathcal{N}_2$ to get the corresponding RAW image and the shared feature vector $l^{\prime}$. We then composite the rendered object with the RAW image using the mask $M$:
\vspace{-3pt}
\begin{eqnarray}
I_R, l &=& \mathcal{N}_2(I_{\mathrm{JPEG}}),\\
 l^{\prime} &=& f(l),\\
I_r &=& M \odot R + (1-M)\odot I_R,
\end{eqnarray}
where $\odot$ denotes the Hadamard product and $I_{\mathrm{JPEG}}$ is the input JPEG photo. Subsequently, we pass $I_r$ and $l^{\prime}$ to $\mathcal{N}_1$ and obtain the predicted JPEG image $I_{pred}$ :
\vspace{-3pt}
\begin{eqnarray}
I_{pred} = \mathcal{N}_1(I_r, l^{\prime}).
\end{eqnarray}
The final composited JPEG image is computed as:
\vspace{-3pt}
\begin{eqnarray}
I_{com} = M \odot I_{pred} + (1-M)\odot I_{\mathrm{JPEG}}.
\end{eqnarray}

\vspace{-5pt}
\section{Experiments}
\vspace{-4pt}
In this section, we extensively evaluate our image translation system. As the object compositing largely relies on the quality of image translation, we first focus our evaluation on various alternative network configurations in Sec.~\ref{sec:exp_img_translation}. The system is further qualitatively validated on compositing results through comparisons to alternative approaches and by conducting user studies in Sec.~\ref{sec:exp_compositing}.

\vspace{-2pt}
\subsection{Data Collection}
\vspace{-2pt}
To train the coupled networks $\mathcal{N}_1$ and $\mathcal{N}_2$, we manually collected 683 RAW-JPEG pairs using a Sony $\alpha$-5100 camera. All photos were acquired with the camera set to auto mode, which results in various color transformation pipelines depending on the scene. Our dataset contains various kinds of scenes including outdoor, indoor, landscape, and portrait. Some examples are shown in the Supplementary Material. We additionally utilize the Canon 5D Mark III dataset from \cite{nam2017modelling}, which contains 645 RAW-JPEG image pairs of various scenes. Although two additional datasets are provided in \cite{nam2017modelling}, we do not use them because we lack access to those camera models for color calibration.

To further diversify the training data, we augment each of the two datasets by simulating various simple pipelines on the RAW images, specifically by applying random RGB rescalings, saturation level adjustments, and a gamma correction from a set of ten common samples. Details on the data augmentation are given in the Supplementary Material.

\begin{table*}
	\begin{center}
	\caption{Comparisons with different network configurations. The two values in each cell represent PSNR values for Canon/Sony images. Cycle(JPEG) denotes results where we feed the output of $\mathcal{N}_2$ to $\mathcal{N}_1$ and get the predicted JPEG images.  Bold text indicates the best performance.}
	\vspace{-7pt}
		\label{tbl:multi}
		\begin{tabular}{lccccccc}
			\toprule
			Sharing-schema & $L_1$ & $L_2$ &Max Pool & Avg Pool& RAW$\to$JPEG &JPEG$\to$RAW& Cycle(JPEG)\\
			\midrule
			\multirow{2}{*}{No Sharing~\cite{nam2017modelling}} &\checkmark&&&&26.29/24.25&\textbf{34.92}/32.42&24.49/25.33\\
			&\checkmark&\checkmark&&&26.03/24.00&34.40/32.39&31.72/32.37\\
			\midrule
			\multirow{4}{*}{Sharing-Conv2}	
			&\checkmark&&\checkmark&&28.65/25.26&34.02/32.42&26.57/27.94\\
			&\checkmark&&&\checkmark&28.80/25.44&34.05/32.57&26.74/28.06\\
			&\checkmark&\checkmark&\checkmark&&28.46/24.99&34.09/32.41&31.70/33.06\\
			&\checkmark&\checkmark&&\checkmark&28.64/25.09&34.20/32.30&31.81/33.09\\
			\midrule
			\multirow{4}{*}{Sharing-Conv1}
			&\checkmark&&\checkmark&&30.84/26.07&34.36/32.59&26.88/28.18\\
			&\checkmark&&&\checkmark&\textbf{31.07/26.16}&34.34/\textbf{32.60}&27.24/28.35\\
			&\checkmark&\checkmark&\checkmark&&30.73/25.75&34.35/32.48&31.84/33.38\\
			&\checkmark&\checkmark&&\checkmark&30.83/25.98&34.16/32.55&\textbf{32.06/33.64}\\
			\bottomrule
		\end{tabular}
	\end{center}
	\vspace{-15pt}
\end{table*}
\vspace{-2pt}
\subsection{Experimental Settings}
\vspace{-2pt}
We first calibrate demosaiced RAW images from both datasets into the canonical color space using the estimated transformation function described in Sec.~\ref{sec:canonical}. For the Sony $\alpha$-5100 dataset that we collected, we set aside $50$ images for testing, use $80\%$ of the remaining images for training, and take the other $20\%$ for validation. For the Canon 5D Mark III dataset, we use the same configuration as in \cite{nam2017modelling}, where the ratio between training and validation is $4:1$, excluding $50$ images for testing. Note that none of the images generated using a simulated pipeline were used as test images. Considering the relatively small size of the datasets that we have, we further augment the data during training. Each RAW-JPEG pair is first randomly left-right flipped or up-down flipped with 0.5 probability for each. Then we crop the image with randomly generated square bounding boxes, which are obtained by first randomly choosing its upper-left corner location, and then randomly selecting the box length, from 128px to the maximum length without extending beyond the image. The crops are resized to 256px$\times$256px to facilitate batch training. In testing, the whole image can be fed into the networks, which can accept input images of arbitrary size.

Our networks are implemented in PyTorch and trained with the Adam optimizer \cite{kingma2014adam}. For a mini batch, we randomly select 8 images with half from the Sony $\alpha$-5100 dataset and the other half from the Canon 5D Mark III dataset. The learning rate is set to $1\times10^{-3}$ for both $\mathcal{N}_1$ and $\mathcal{N}_2$. The hyperparameter $\lambda$ is set to 1 in all experiments.

\vspace{-2pt}
\subsection{Image Translation}
\label{sec:exp_img_translation}
\subsubsection{Different Network Configurations}
\begin{figure*}
	\begin{minipage}[t] {0.24\textwidth}
		\centering
		\includegraphics[width=\textwidth]{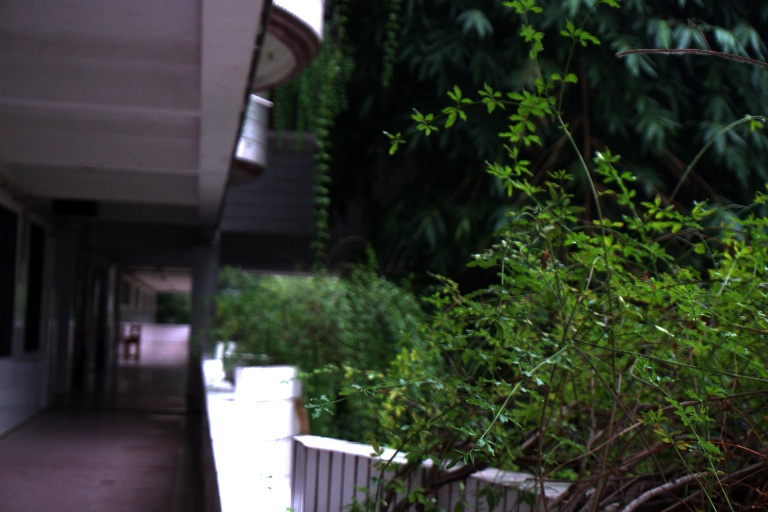}
		{Input}
	\end{minipage}
	\begin{minipage}[t] {0.24\textwidth}
		\centering
		\includegraphics[width=\textwidth]{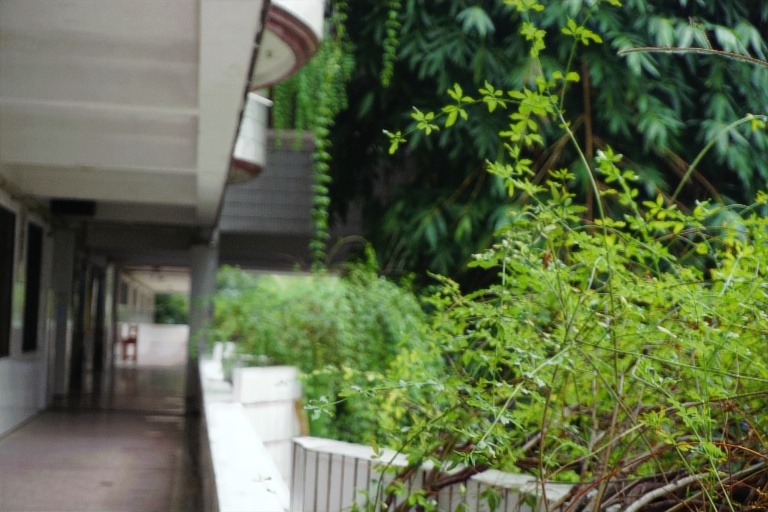}
		{Cycle + Feature Sharing}
	\end{minipage}
	\begin{minipage}[t] {0.24\textwidth}
		\centering
		\includegraphics[width=\textwidth]{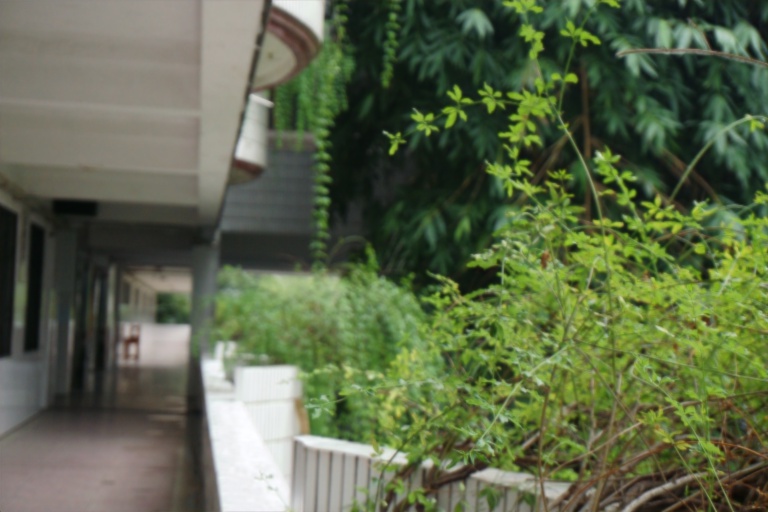}
		{Feature Sharing}
	\end{minipage}
	\begin{minipage}[t] {0.24\textwidth}
		\centering
		\includegraphics[width=\textwidth]{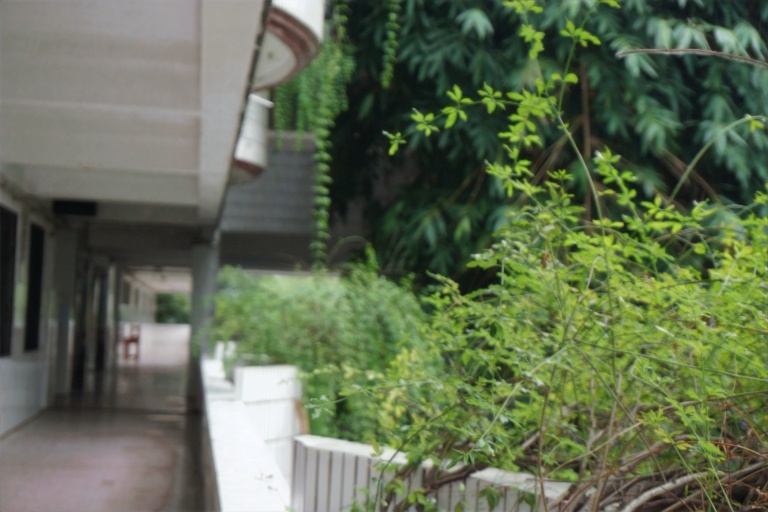}
		{Baseline}
	\end{minipage}
	\begin{minipage}[t] {0.24\textwidth}
		\centering
		\includegraphics[width=\textwidth]{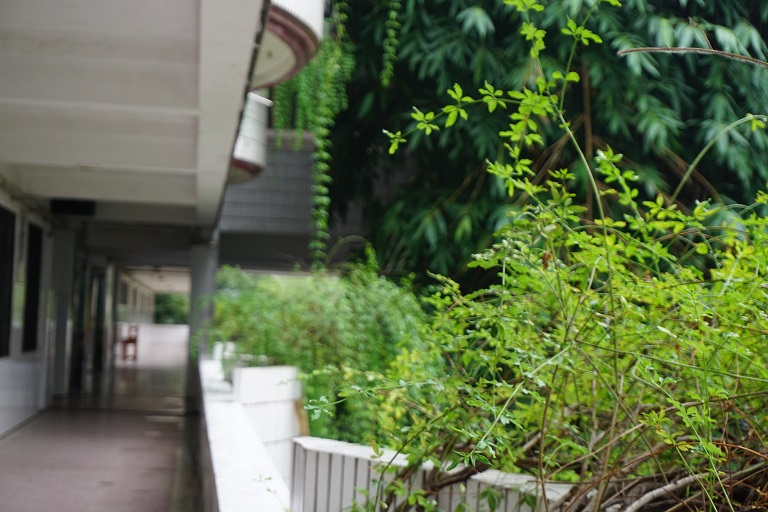}
		{Ground Truth}
	\end{minipage}
	{}
	\begin{minipage}[t] {0.755\textwidth}
		\centering
		\includegraphics[width=\textwidth]{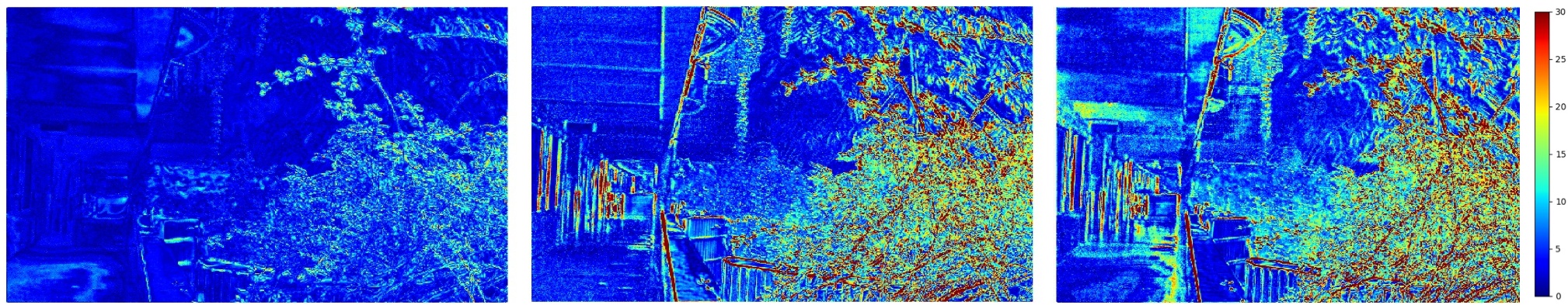}
		\begin{minipage}[t]{0.32\textwidth}
		\centering
		{Cycle + Feature Sharing}
		\end{minipage}
		\begin{minipage}[t]{0.32\textwidth}
		\centering
		{Feature Sharing}
		\end{minipage}
		\begin{minipage}[t]{0.32\textwidth}
		\centering
		{Baseline}
		\end{minipage}
				
	\end{minipage}
	\vspace{-7pt}
	\caption{The first row shows predictions using different network configurations. The corresponding error maps are displayed in the second row. The baseline method utilizes no feature sharing or cycle consistency.}
	\label{fig:error_map}
\end{figure*}

\vspace{-2pt}
We compare our model with its variants  that employ other loss functions, feature sharing schema, and base networks. Due to the wide use in the literature~\cite{nam2017modelling}, the Peak Signal-to-Noise Ratio (PSNR) with respect to test sets of both the Canon 5D Mark III and Sony $\alpha$-5100  is used as metric. The results, shown in Table~\ref{tbl:multi}, Table~\ref{table:img_translation_others}, and Figure~\ref{fig:error_map}, are discussed in the following. We also measure performance using CIE Delta E 2000~\cite{brainard2003color} and find the results are consistent with that from PSNR, as shown in the supplement.

\vspace{-12pt}
\paragraph{Cycle Consistency Constraint}
As shown in Table~\ref{tbl:multi}, including the cycle consistency constraint ($L_2$) leads to better results for both datasets, regardless of whether feature sharing is enabled. Note that the performance of RAW-to-JPEG prediction becomes slightly worse. We hypothesize that, as the two networks are trained jointly, they would implicitly cooperate with each other to achieve a smaller loss on the joint prediction task (JPEG-to-RAW-to-JPEG) at the cost of degrading performance on a single task (JPEG-to-RAW or RAW-to-JPEG).
\vspace{-12pt}
\paragraph{Shared Features}
We observe that JPEG prediction is better with feature sharing than without it. The RAW-to-JPEG prediction improves from 26.03/24.00 to 30.83/25.98 in terms of PSNR on the Canon/Sony Datasets, and the cycle JPEG prediction performance also increases on the two datasets, by 0.34/1.27. Sharing global features related to the in-camera color processing pipeline effectively removes ambiguity in JPEG prediction and generates better results.
\vspace{-12pt}
\paragraph{Sharing Methods}
We examine the use of different hidden layers in $\mathcal{N}_2$ and functions $f$ for feature sharing. As indicated by the results in Table~\ref{tbl:multi}, the performance becomes worse by taking the feature map of deeper layers. This result is expected, as deeper features provide a more semantic representation of RAW images and are less reflective of the JPEG coloring properties. We thus use the feature map of the first convolutional layer in our system. Among variants of function $f$, we find average pooling slightly outperforms max pooling.

\begin{table}
		\caption{Comparisons with different base networks. The two values in each cell represent PSNR values for Canon/Sony images.}
		\label{table:img_translation_others}
		\vspace{-9pt}
		\begin{adjustbox}{width=0.48\textwidth}
		\begin{tabular}{lccc}
			\toprule
			Network & RAW$\to$JPEG &JPEG$\to$RAW& Cycle(JPEG)\\
			\midrule
			MLP & 23.44/21.90&31.97/31.62&\textbf{34.93/35.06}\\
			SRCNN~\cite{dong2014learning}&26.17/23.21&33.34/\textbf{32.64}&31.82/32.76\\
			UNet~\cite{ronneberger2015u} &26.13/23.56&32.70/32.37&31.72/32.83\\
			Multi&\textbf{30.83/25.98}&\textbf{34.16}/32.55&32.06/33.64\\
			\bottomrule
		\end{tabular}
		\end{adjustbox}
\end{table}
\vspace{-7pt}
\if(0)
\begin{table*}
	\begin{center}
		\caption{Comparisons with different base networks. The two values in each cell represent PSNR values for Canon/Sony images.}
		\vspace{-9pt}
		\label{table:img_translation_others}
		\begin{tabular}{lcccccc}
			\toprule
			Network & $L_1$ & $L_2$ & Sharing&RAW$\to$JPEG &JPEG$\to$RAW& Cycle(JPEG)\\
			\midrule
			\multirow{4}{*}{MLP} &\checkmark&&&22.78/21.84&31.72/31.53&28.51/28.60\\
			&\checkmark&\checkmark&&22.61/21.67&31.75/31.42&38.61/39.32\\
			&\checkmark&&\checkmark&23.87/22.21&31.98/31.72&26.62/28.20\\
			&\checkmark&\checkmark&\checkmark&23.44/21.90&31.97/31.62&34.93/35.06\\
			\midrule
			\multirow{4}{*}{SRCNN}&\checkmark&&&&&\\
			&\checkmark&\checkmark&&&&\\
			&\checkmark&&\checkmark&26.93/23.89&33.06/32.57&22.54/24.58\\
			&\checkmark&\checkmark&\checkmark&26.17/23.21&33.34/32.64&31.82/32.76\\
			\midrule
			\multirow{4}{*}{UNet} &\checkmark&&&25.61/24.35&33.17/32.75&22.73/24.72\\
			&\checkmark&\checkmark&&24.71/23.74&33.35/32.06&32.06/33.32\\
			&\checkmark&&\checkmark&26.70/24.07&32.91/32.38&22.91/24.59\\
			&\checkmark&\checkmark&\checkmark&26.13/23.56&32.70/32.37&31.72/32.83\\
			\midrule
			\multirow{4}{*}{Multi} &\checkmark&&&26.41/24.29&34.94/32.64&24.55/25.64\\
			&\checkmark&\checkmark&&25.98/24.08&34.64/32.43&31.15/31.73\\
			&\checkmark&&\checkmark&30.14/25.96&34.60/32.73&26.27/27.72\\
			&\checkmark&\checkmark&\checkmark&29.97/25.76&34.64/32.69&31.76/32.75\\
			\bottomrule
		\end{tabular}
	\end{center}
\end{table*}
\fi
\begin{table}
		\caption{PSNR results on unknown cameras.}
		\label{tbl:unknown_cameras}
		\vspace{-9pt}
		\begin{tabular}{lccc}
			\toprule
			Datasets & RAW$\to$JPEG & JPEG$\to$RAW & Cycle\\
			\midrule
			Canon 60D & 27.45 & 30.50 & 32.71\\
			Sony NEX 7& 26.93 & 31.58 & 32.74\\
			\bottomrule
		\end{tabular}
		\vspace{-5pt}
\end{table}

\vspace{-8pt}
\paragraph{Base Networks}
We also experiment with different image translation models as our base network. Specifically, we consider four different types of neural networks: Multi-layer Perceptron (MLP), SRCNN~\cite{dong2014learning}, UNet~\cite{ronneberger2015u}, and Multi-Scale Learnable Histograms \cite{nam2017modelling}. The configurations for the different neural networks are given in the supplement, and the networks are trained using cycle consistency and feature sharing. From the results listed in Table~\ref{table:img_translation_others}, it can be seen that the Multi-Scale Learnable Histogram performs best on both RAW-to-JPEG and JPEG-to-RAW. This was also found to be the case with other network settings, whose results are also provided in the supplement. Although the other networks give slightly better cycle results (RAW-to-JPEG-to-RAW), we found that their lower performance on RAW-to-JPEG leads to poorer compositing results.

\begin{figure*}
	\begin{center}
		\begin{minipage}[t] {0.24\textwidth}
			\centering
			\includegraphics[width=\textwidth]{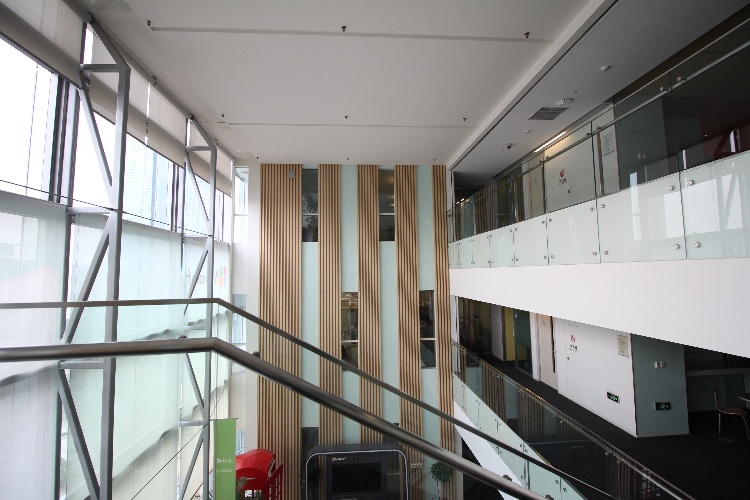}
		\end{minipage}
		\begin{minipage}[t] {0.24\textwidth}
			\centering
			\includegraphics[width=\textwidth]{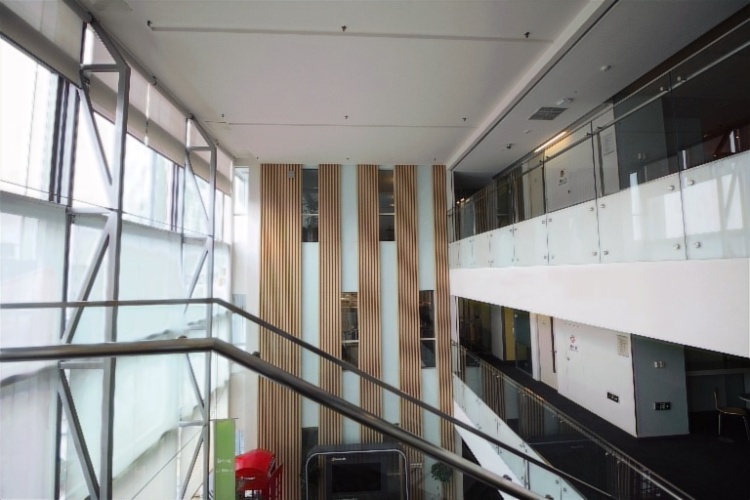}
		\end{minipage}
		\begin{minipage}[t] {0.24\textwidth}
			\centering
			\includegraphics[width=\textwidth]{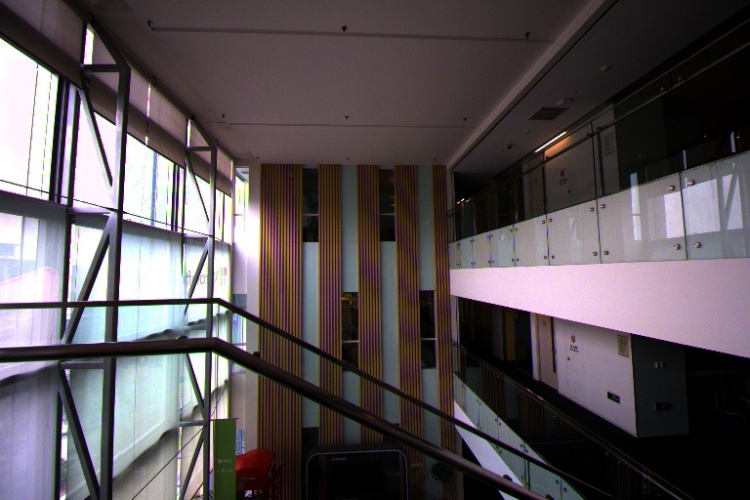}
		\end{minipage}
		\begin{minipage}[t] {0.24\textwidth}
			\centering
			\includegraphics[width=\textwidth]{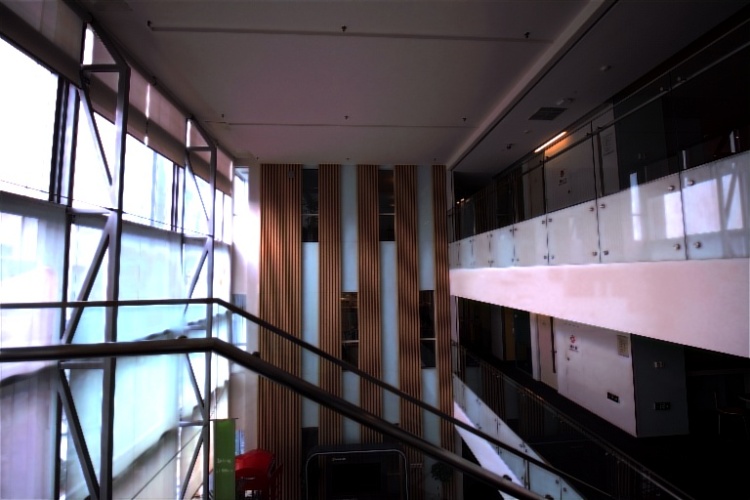}
		\end{minipage}
		\begin{minipage}[t]{\textwidth}
		\centering
		\begin{minipage}[t] {0.24\textwidth}
			\centering
			\includegraphics[width=\textwidth]{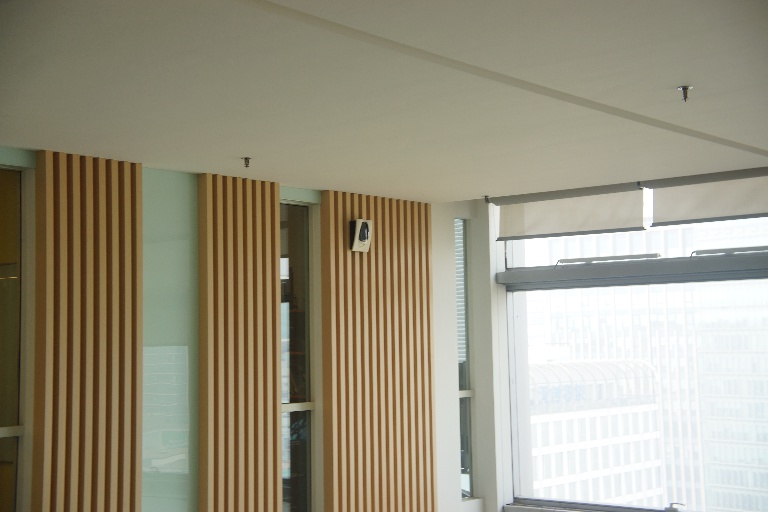}
			{Input JPEG}
		\end{minipage}
		\begin{minipage}[t] {0.24\textwidth}
			\centering
			\includegraphics[width=\textwidth]{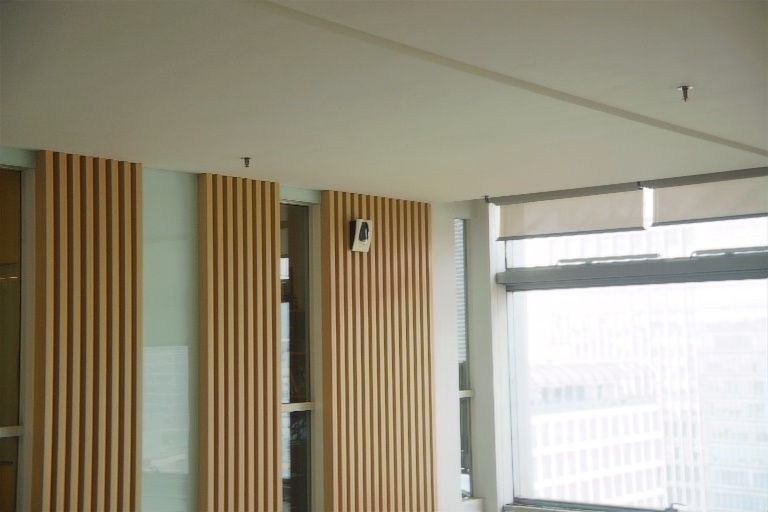}
			{Predicted JPEG}
		\end{minipage}
		\begin{minipage}[t] {0.24\textwidth}
			\centering
			\includegraphics[width=\textwidth]{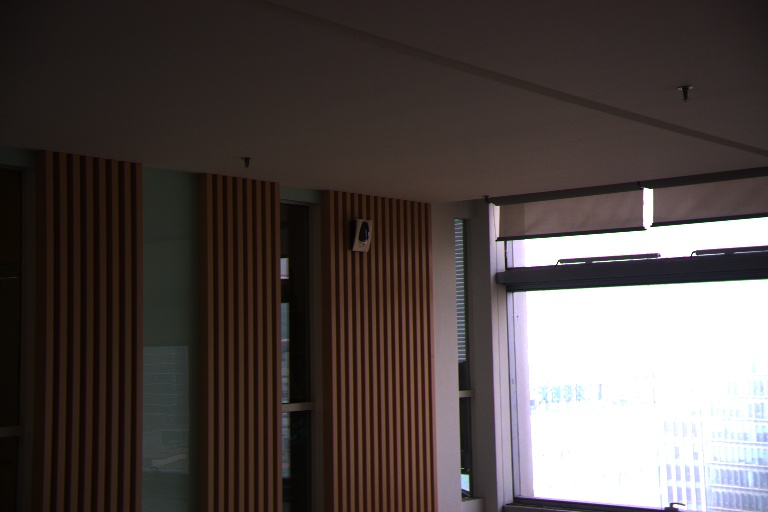}
			{Calibrated RAW}
		\end{minipage}
		\begin{minipage}[t] {0.24\textwidth}
			\centering
			\includegraphics[width=\textwidth]{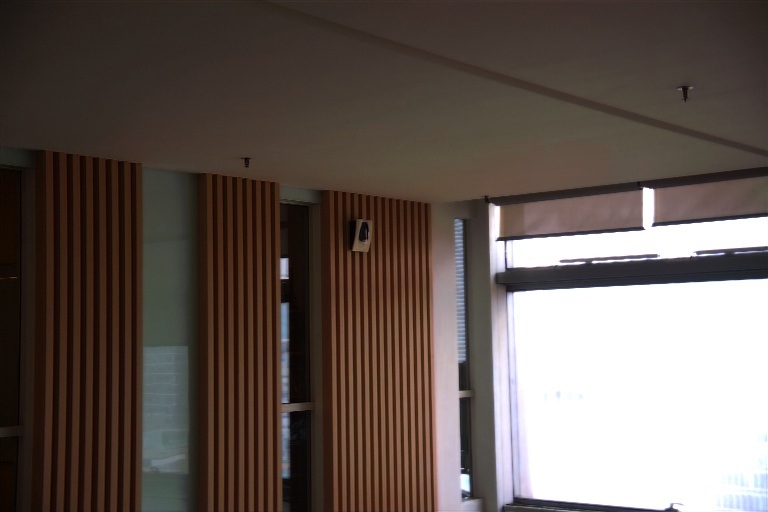}
			{Predicted RAW}
		\end{minipage}
		\end{minipage}
		\vspace{-7pt}
		\caption{The first/second row shows the predictions for a photo from Canon 60D/Sony NEX7.}
		\label{fig:unknown_cameras}
	\end{center}
\end{figure*}
\begin{figure*}
	\centering
	\begin{minipage}[t] {0.19\textwidth}
		\centering
		\includegraphics[width=\textwidth]{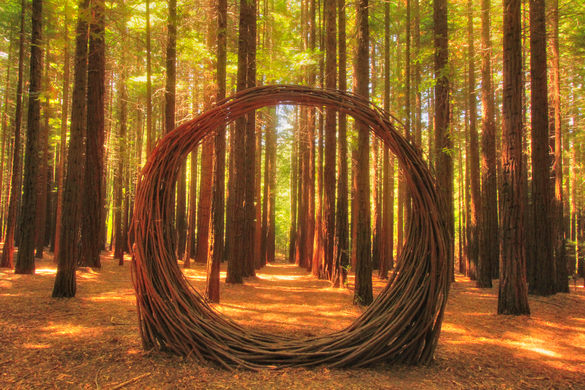}
		{(a)}
	\end{minipage}
	\begin{minipage}[t] {0.19\textwidth}
		\centering
		\includegraphics[width=\textwidth]{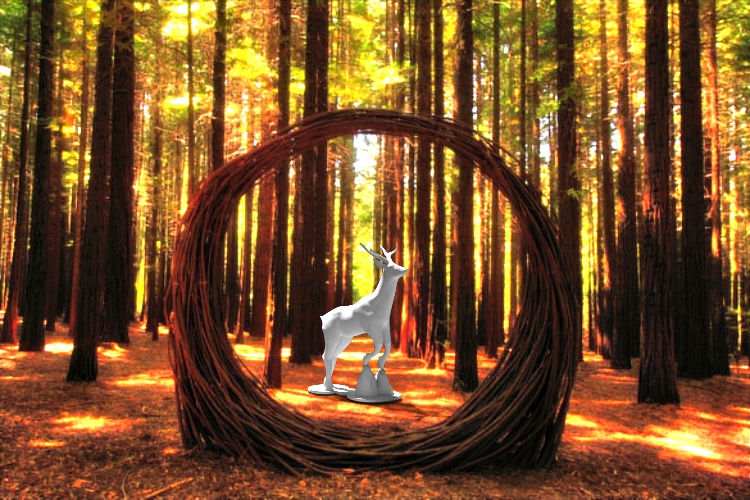}
		{(b)}
	\end{minipage}
	\begin{minipage}[t] {0.19\textwidth}
		\centering
		\includegraphics[width=\textwidth]{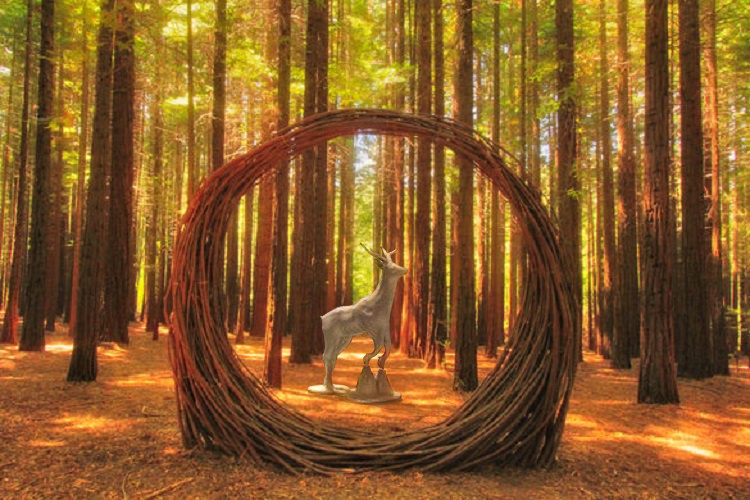}
		{(c)}
	\end{minipage}
	\begin{minipage}[t] {0.19\textwidth}
		\centering
		\includegraphics[width=\textwidth]{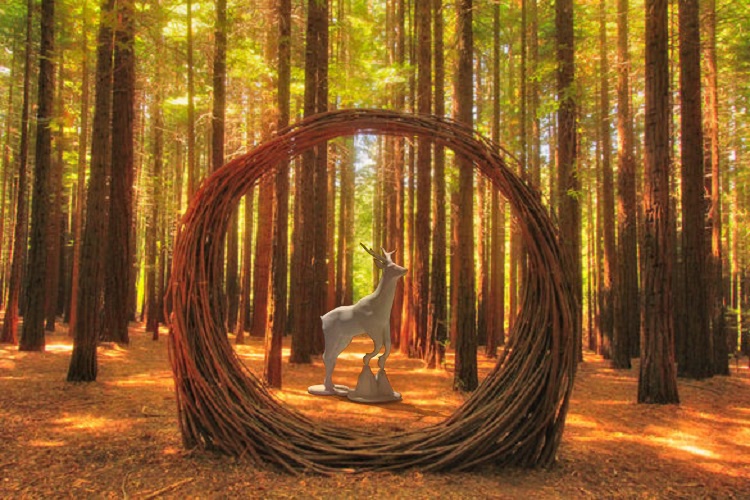}
		{(d)}
	\end{minipage}
	\begin{minipage}[t] {0.19\textwidth}
		\centering
		\includegraphics[width=\textwidth]{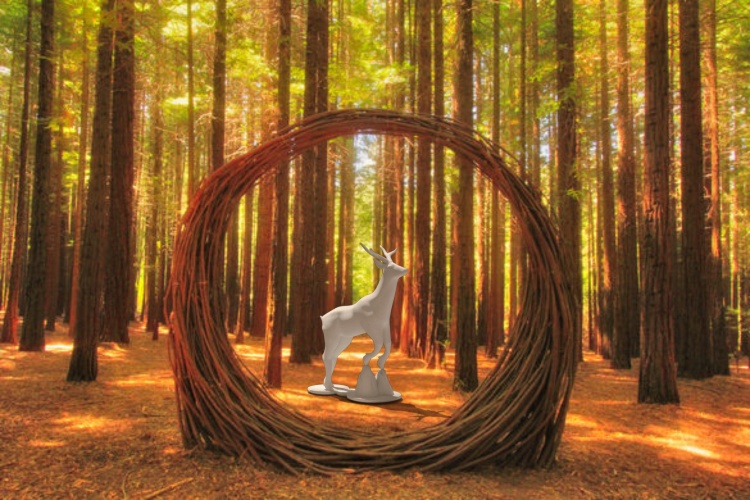}
		{(e)}
	\end{minipage}
	\vspace{-7pt}
	\caption{ Comparisons to related image translation methods, where the deer is the composited object. (a) Input JPEG. (b) Blended RAW with linear scaling. (c) Style transfer by \cite{gatys2016image}. (d) Color transfer by \cite{luan2017deep}. (e) Our method. [Please zoom-in.]}
	\label{fig:transfer_imgs}
\end{figure*}
\begin{figure*}
	\centering
	\begin{minipage}[t] {0.19\textwidth}
		\centering
		\includegraphics[width=\textwidth]{images/style_transfer/input.jpg}
		{(a)}
	\end{minipage}
	\begin{minipage}[t] {0.19\textwidth}
		\centering
		\includegraphics[width=\textwidth]{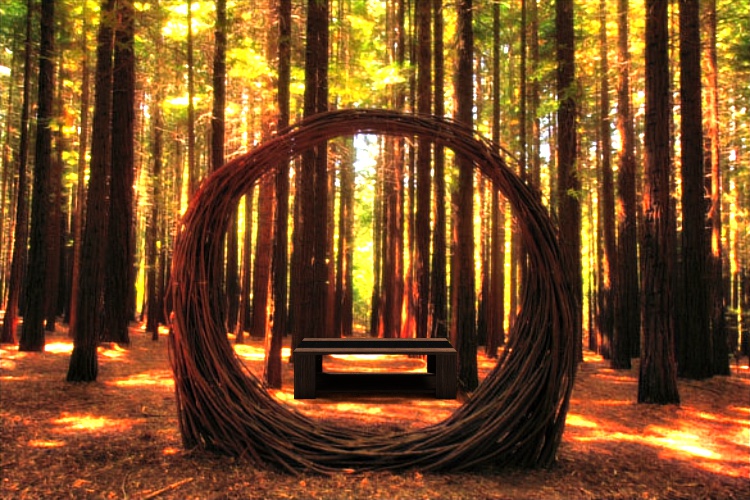}
		{(b)}
	\end{minipage}
	\begin{minipage}[t] {0.19\textwidth}
		\centering
		\includegraphics[width=\textwidth]{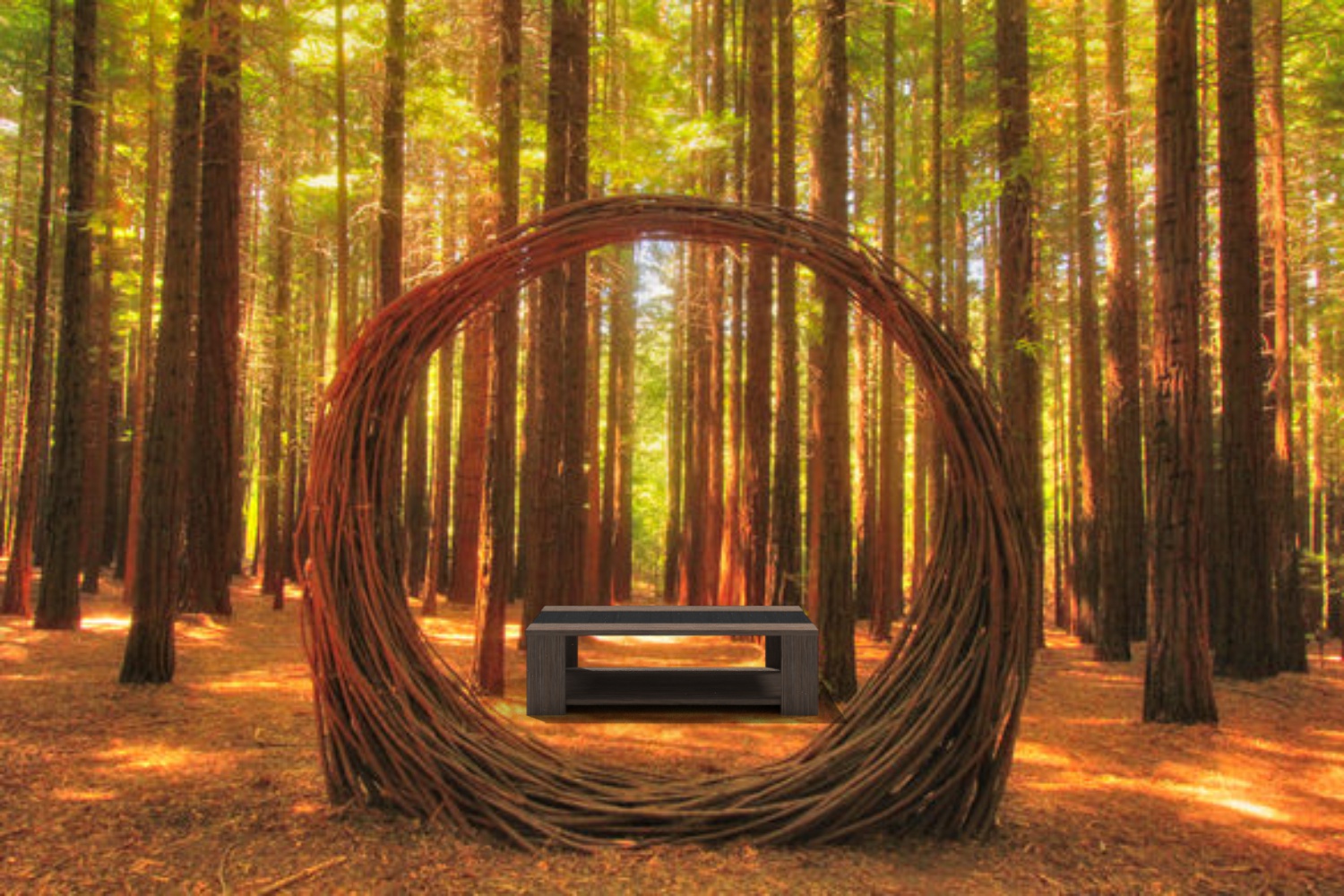}
		{(c)}
	\end{minipage}
	\begin{minipage}[t] {0.19\textwidth}
		\centering
		\includegraphics[width=\textwidth]{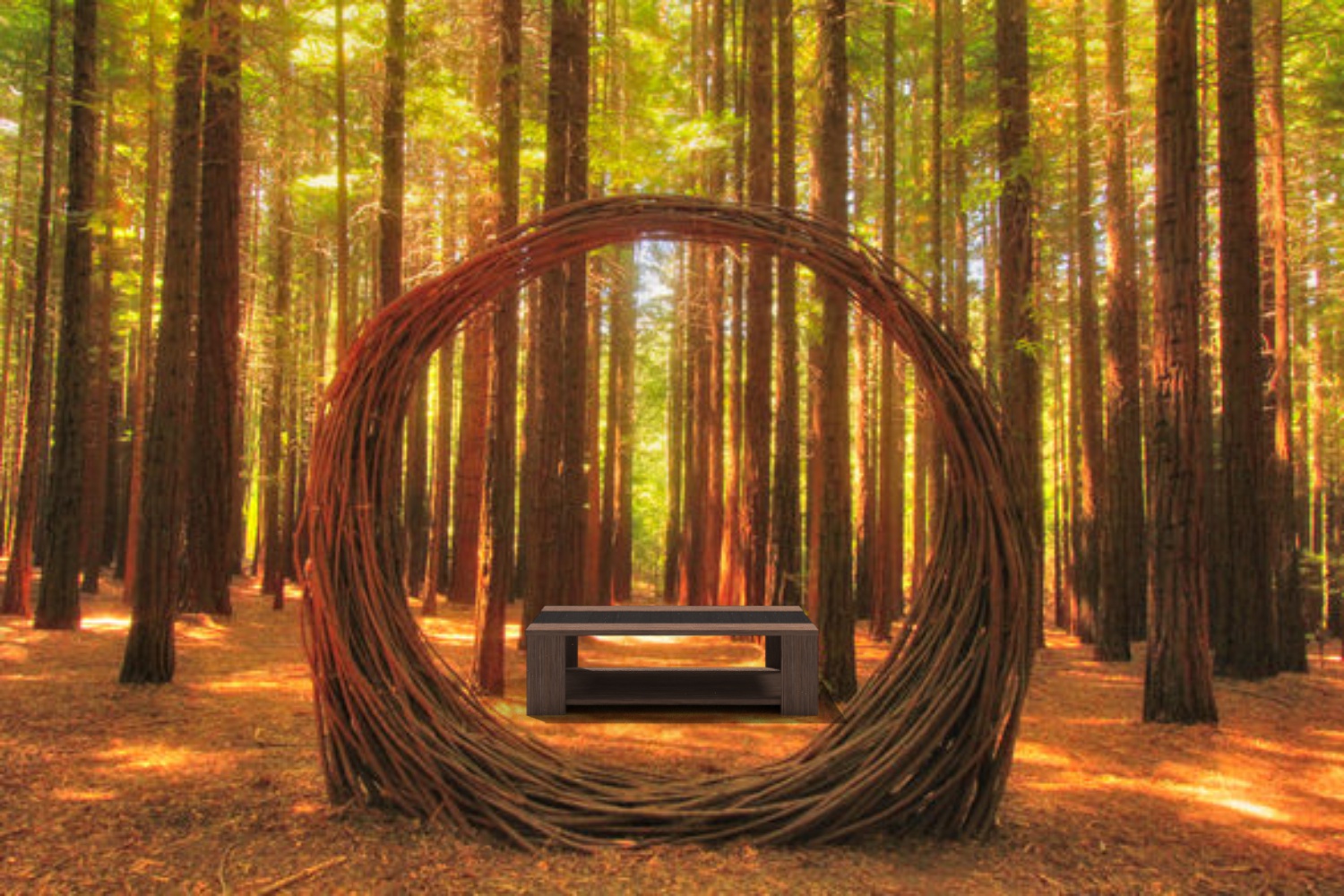}
		{(d)}
	\end{minipage}
	\begin{minipage}[t] {0.19\textwidth}
		\centering
		\includegraphics[width=\textwidth]{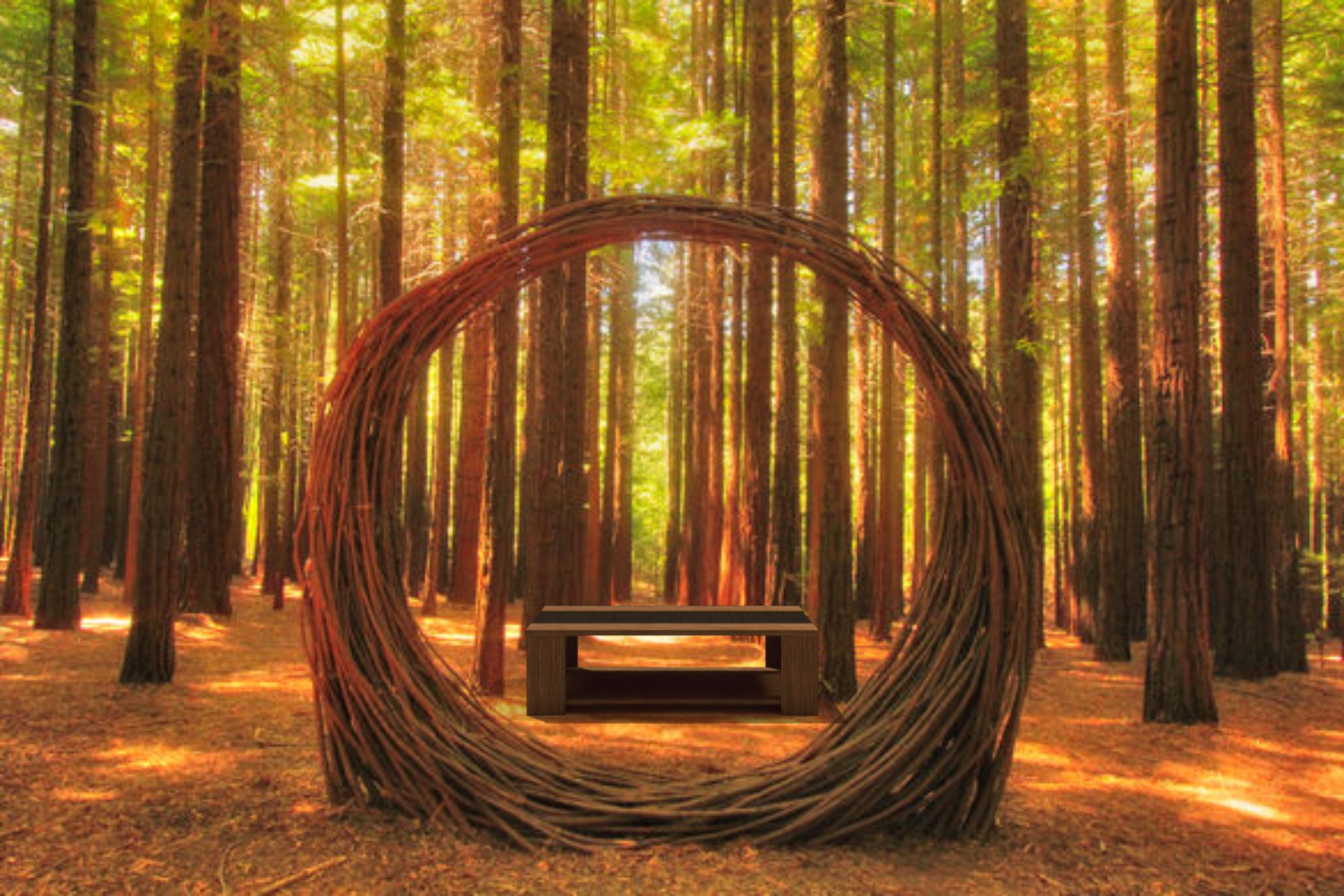}
		{(e)}
	\end{minipage}
	\vspace{-8pt}
	\caption{Comparison to estimated white balance, where the desk is the composited object.  (a) Input JPEG. (b) Blended RAW with linear scaling. (c) Gamma correction. (d) White balance and gamma correction. (e) Our method. [Please zoom-in.]}
	\vspace{-10pt}
	\label{fig:wb_compare}
\end{figure*}

\vspace{-4pt}
\subsubsection{Unknown Cameras}
As the feature sharing schema is designed to extract features that represent the JPEG rendering characteristics of the imaging pipelines, our model should be able to generalize to unknown cameras. To verify this, we first train the model on images from just a single camera (Canon 5D Mark III, specifically) and test it on images from the training camera and another camera (Sony $\alpha$-5100, specifically). The model achieves a 31.10/35.98 (Raw$\to$JPEG/Cycle) PSNR on the same camera and a 24.69/36.30 PSNR on the Sony $\alpha$-5100 camera, exhibiting a moderate level of generalization ability from only a single training camera. 

As is the case for other convolutional networks, more generalizable CNN features can be learned by providing a broader distribution of training data, i.e. from multiple cameras. We thus additionally train the model on images from two cameras (Canon 5D Mark III and Sony $\alpha$-5100) and collect 50 other RAW-JPEG pairs using a Canon 60D and Sony NEX for testing. Although these models are from the same company as the training cameras, we observed differences in the color transformation pipelines from images taken of the same scenes (see the supplement for examples). We directly test the trained model on these datasets without finetuning. The prediction results are shown in  Table~\ref{tbl:unknown_cameras}. Though the PSNR values are slightly lower than those in Table~\ref{tbl:multi}, they are at a similar level. Examples of JPEG inputs, predicted RAW images, ground truth calibrated RAW images, and predicted JPEG images (after JPEG-to-RAW and RAW-to-JPEG) are shown in Figure~\ref{fig:unknown_cameras}.

\begin{table}
	\begin{center}
	{
		\caption{User study results, in terms of selection percentage}
		\label{table:user_study}
		\vspace{-8pt}
		\begin{adjustbox}{width=0.42\textwidth}
		\begin{tabular}{lccc}
			\toprule
			& Gamma & Gamma+WB & Our method \\
			\hline
			Max & 58.3\% & 58.3\% &\textbf{91.7\%} \\
			Min & 16.7\% & 4.2\% & \textbf{29.2\%}\\
			Average &24.6\% & 38.0\% & \textbf{58.5\%}\\
			\bottomrule
		\end{tabular}
		\end{adjustbox}
	}
	\end{center}
	\vspace{-20pt}
\end{table}
\vspace{-2pt}
\subsection{Compositing Objects}
\label{sec:exp_compositing}
\subsubsection{External Comparison}
\vspace{-3pt}
An alternative approach to our problem is to apply related image-to-image translation techniques such as style transfer \cite{gatys2016image} or color transfer \cite{luan2017deep}. Specifically, these methods could be used to transfer style or color from the JPEG photo to the virtual object prior to compositing. The results are presented in Figure~\ref{fig:transfer_imgs}. It can be seen that neither of these approaches are suitable for this problem. The style transfer method~\cite{gatys2016image} copies textural properties of the JPEG photo to the virtual object, producing an unnatural-looking result that is inconsistent with the object's white plaster material. Even when only color is transferred, as with~\cite{luan2017deep}, the transferred colors reflect the intrinsic colors of objects in the scene in addition to the JPEG color processing. By extracting and applying the color processing from the JPEG input, our method produces the most satisfactory results. 

Another technique we compare to is the use of white balance estimation with gamma correction, where the inverse of the white balance (estimated using the method of \cite{hu2017}) is applied to the virtual object and is followed by a gamma correction of 2.2.  Figure~\ref{fig:wb_compare} shows a comparison with our method. One can observe that white balance plus gamma does not adequately approximate the downloaded photo's color processing, which includes a boost in saturation. By contrast, our neural network model is powerful enough to capture such color transformations, as seen by the more saturated composited object. 

\begin{figure}
	\centering
	\includegraphics[height=2.7cm]{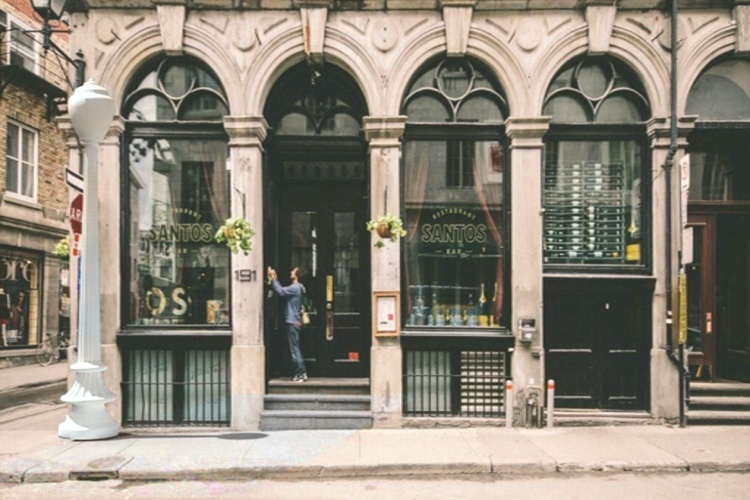}
	\includegraphics[height=2.7cm]{images/pred_jpeg.jpg}
	\vspace{-8pt}
\caption{Comparison with DIH~\cite{tsai2017deep}. Left: result from DIH. Right: result from our method. [Please zoom-in.]}
	\label{fig:dih}
	\vspace{-10pt}
\end{figure}

Another alternative approach is to harmonize the foreground object and background through Deep Image Harmonization (DIH)~\cite{tsai2017deep}. We present the results in Figure~\ref{fig:dih}. It can be seen that although DIH produces a visually pleasing result where the object color looks aesthetically compatible with the surroundings, this is not the same as being photometrically correct with respect to the imaging pipeline. In this particular case, it can be seen that the inserted object for DIH exhibits color variations that are inconsistent with the actual object. 

\vspace{-8pt}
\subsubsection{User Study}
We conducted a user study to evaluate the visual quality of our compositing results. Virtual objects were composited into 24 images for this study. The images were all downloaded from the web and were taken by unknown cameras, with some having an Instagram-style appearance. For comparison, the objects are also composited into the images using a default gamma correction of 2.2 or gamma correction with an estimated white balance~\cite{hu2017}. For each image, our result and the comparisons are shown in random order. The users are asked to select which of the three images appears more natural. A total of 25 users participated in this study.

The results are presented in Table~\ref{table:user_study}, which shows statistics on the percentage of times a method's result was selected. Max/Min are the maximum and minimum percentages from among all the users. It is seen that the users clearly prefer the results of our method over compositing using gamma correction (with/without) white balance. Images for the user study are provided in the supplement.

\begin{figure}
	\centering
	\begin{minipage}[t] {\linewidth}
		\centering
		\includegraphics[width=0.32\linewidth]{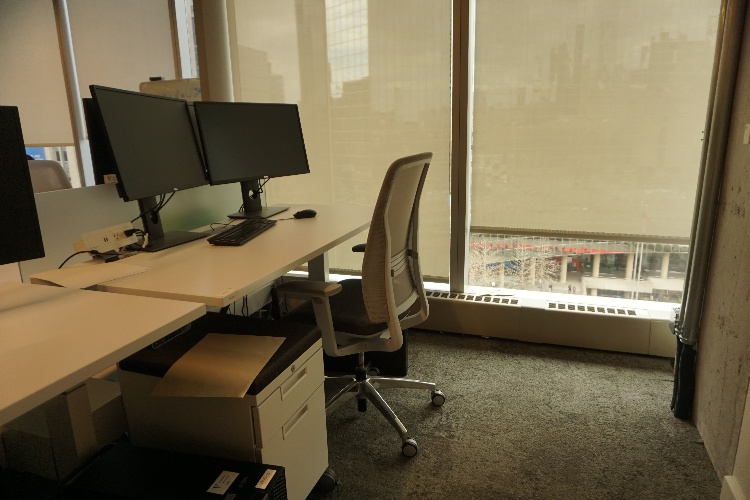}
		\includegraphics[width=0.32\linewidth]{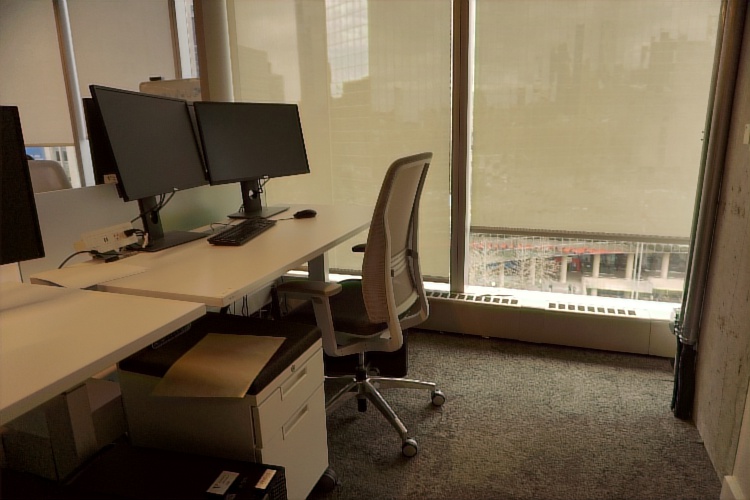}
		\includegraphics[width=0.32\linewidth]{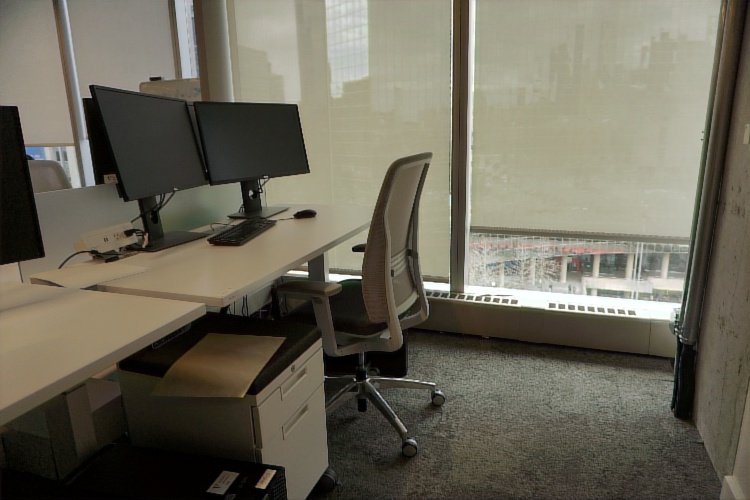}
	\end{minipage}
	\begin{minipage}[t] {\linewidth}
		\centering
		\includegraphics[width=0.32\linewidth]{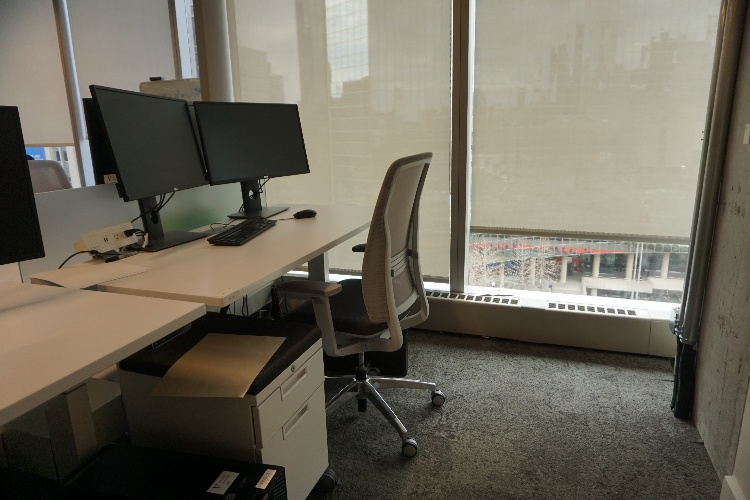}
		\includegraphics[width=0.32\linewidth]{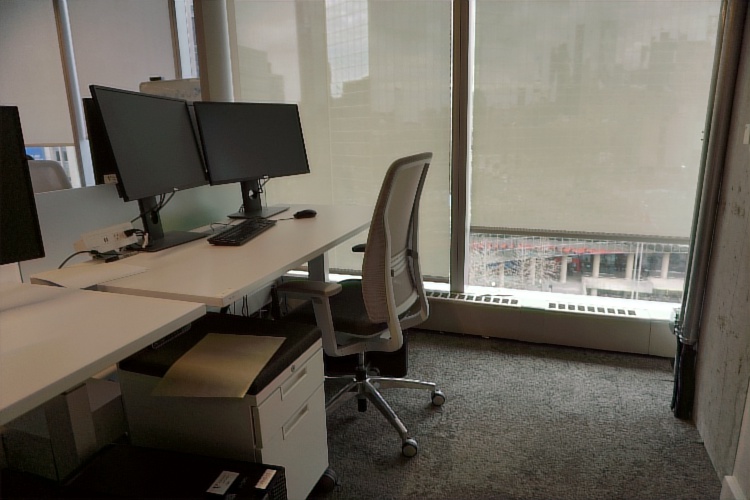}
		\includegraphics[width=0.32\linewidth]{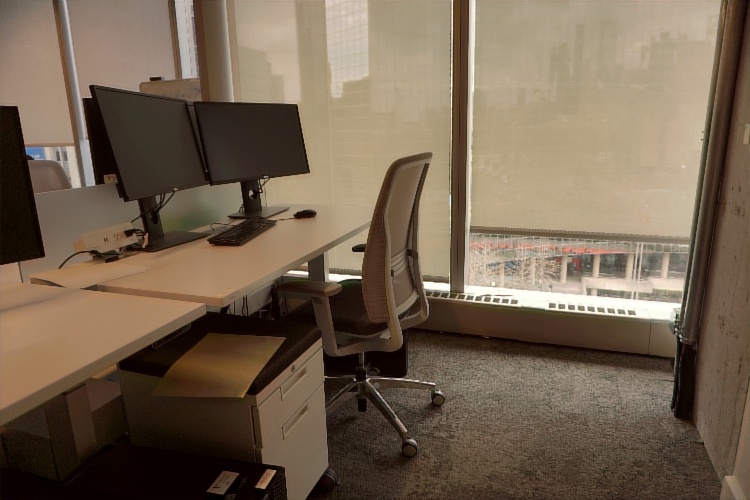}
	\end{minipage}
	\vspace{-15pt}
	\caption{Analysis on the shared features. The first column is the input JPEG; the second column is the network prediction following the standard procedure; the last column is the prediction by swapping shared features. [Please zoom-in.]}
	\vspace{-10pt}
	\label{fig:shared_feature}
\end{figure}
\subsection{Analysis}
To further analyze the characteristics of shared features in the proposed pipeline, we swap the shared features for two different images that capture the same scene but undergo different color pipelines. Specifically, we first capture two photos of the same scene using different camera settings, feed the JPEG to the network $\mathcal{N}_2$ separately and obtain the corresponding RAW and shared features for each photo, then we swap the shared features and use network $\mathcal{N}_1$ to predict a new JPEG. Results are shown in Figure~\ref{fig:shared_feature}. It can be seen that, by swapping the shared features, the colors in the predicted images are also swapped. On the other hand, using the original shared features leads to predictions consistent with the input. This demonstrates that the shared features actually capture the color characteristics of the input JPEG.
\section{Conclusion and Future Work}
\vspace{-4pt}
We presented an object compositing system that estimates the color transformation in the imaging pipeline of the target photo. To solve for this transformation from a single image, we propose a dual learning approach that is made tractable through the use of shared features from the dual to the primal network. We believe that this strategy could be useful for other problems in which a network needs to infer a particular solution in an inherently one-to-many mapping. 


Our system is designed to model global color transformations. There exist some advanced imaging pipelines that may process certain image regions differently from others, for example, by detecting the sky region in an outdoor photo and making it more blue. How to extend our model to handle spatial variations in color processing would be an interesting direction for future study. Another avenue for further work is to adapt the image translation model with model compression techniques such that it could run on mobile devices with fast inference time.

\section{Acknowledgement}
This work is partially supported by National Basic Research Program of China (973 Program) (grant no. 2015CB352502), NSFC (61573026), BJNSF (L172037), and a grant from Microsoft Research Asia.

{\small
\bibliographystyle{ieee}
\bibliography{egbib}
}

\end{document}